\documentclass[10pt,twocolumn,letterpaper]{article}

\usepackage[pagenumbers]{cvpr} 

\usepackage{pifont}
\usepackage{multirow}
\usepackage{bbm}
\usepackage{mathtools}
\usepackage{float}
\usepackage{caption}
\usepackage{makecell}
\usepackage{algorithm}
\usepackage{algpseudocode}
\usepackage{tikz}
\usetikzlibrary{shapes.geometric}
\usepackage{tikz-network}
\usepackage{tikz-3dplot}
\usepackage{pgfplots}
\usepackage{enumitem}
\DeclareMathOperator*{\argmin}{arg\,min}

\newcommand{\AlgName}{{VTCD}\xspace}
\newcommand{\ConceptRankAlgName}{{CRIS}\xspace}
\DeclareRobustCommand\tikzbluecircle{\tikz \draw[fill=blue] (1ex,1ex) circle (1ex);}
\DeclareRobustCommand\tikzgreensquare{\tikz \draw[fill=green] (0.1,0.1) rectangle (0.35,0.35);}
\DeclareRobustCommand\tikzpurplediamond{\tikz{\draw[fill=purple,rotate=45] (0.1,0.1) rectangle (0.3,0.3);}}
\DeclareRobustCommand\tikzorangetriangle{\tikz \node[isosceles triangle,
	draw,
	rotate=90,
	fill=orange,
        inner sep=0pt,
	minimum size=0.2cm] (T1)at (0,0){};}

\definecolor{cvprblue}{rgb}{0.21,0.49,0.74}
\usepackage[pagebackref,breaklinks,colorlinks,citecolor=cvprblue]{hyperref}

\def\paperID{2544} 
\def\confName{CVPR}
\def\confYear{2024}
\newcommand{\smallsec}[1]{\vspace{0.2em}\noindent\textbf{#1}}

\title{Understanding Video Transformers via Universal Concept Discovery}

\author{Matthew Kowal$^{1,3,4}$\thanks{Work completed during an internship at Toyota Research Institute} \hspace{0.4cm}
Achal Dave$^{3}$ \hspace{0.4cm}
Rares Ambrus$^{3}$ \\
Adrien Gaidon$^{3}$ \hspace{0.4cm}
Konstantinos G. Derpanis$^{1,2,4}$ \hspace{0.4cm}
Pavel Tokmakov$^{3}$ \\ \vspace{0.1cm}
{\normalsize $^{1}$York University, $^{2}$Samsung AI Centre Toronto, $^{3}$Toyota Research Institute, $^{4}$Vector Institute}\vspace{-0.07cm} \\ \vspace{-0.12cm}
{Project page: \href{https://yorkucvil.github.io/VTCD}{yorkucvil.github.io/VTCD}}
}

\begin{document}
\maketitle
\begin{abstract}

This paper studies the problem of concept-based interpretability of transformer representations for videos. Concretely, we seek to explain the decision-making process of video transformers based on high-level, spatiotemporal concepts that are automatically discovered. 
Prior research on concept-based interpretability has concentrated solely on image-level tasks. Comparatively, video models deal with the added temporal dimension, increasing complexity and posing challenges in identifying dynamic concepts over time.
In this work, we systematically address these challenges by introducing the first Video Transformer Concept Discovery (\AlgName) algorithm. To this end, we propose an efficient approach for unsupervised identification of units of video transformer representations - concepts, and ranking their importance to the output of a model. The resulting concepts are highly interpretable, revealing spatio-temporal reasoning mechanisms and object-centric representations in unstructured video models. Performing this analysis jointly over a diverse set of supervised and self-supervised representations, we discover that some of these mechanism are universal in video transformers. Finally, we show that \AlgName can be used for fine-grained action recognition and video object segmentation.
\end{abstract}    
\section{Introduction}
\label{sec:intro}

\begin{figure}
    \centering   
\includegraphics[width=0.5\textwidth]{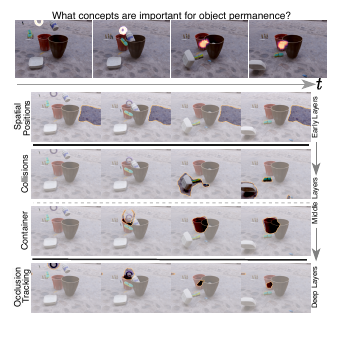}
    \vspace{-40pt}
    \caption{Heatmap predictions of the TCOW model~\cite{van2023tracking} for tracking through occlusions (top), together with concepts discovered by our \AlgName (bottom). 
    We can see that the model encodes positional information in early layers, identifies containers and collision events in mid-layers and tracks through occlusions in late layers.  Only one video is shown, but the discovered concepts are shared between many dataset samples (see \href{https://youtu.be/TVfSyDQAb3I}{video} for full results).
    }
    \vspace{-20pt}
    \label{fig:Teaser}
\end{figure}

Understanding the hidden representations within neural networks is essential for addressing regulatory concerns~\cite{euro2021laying,whitehouse2023president}, preventing harms in deployment~\cite{buolamwini2018gender,hansson2021self}, and can aid innovative model designs~\cite{darcet2023vision}. 
This problem has been studied extensively for images, both for convolutional neural networks (CNNs)~\cite{bau2017network,kim2018interpretability,ghorbani2019towards,fel2023craft} and, more recently, vision transformers (ViTs)~\cite{raghu2021vision,walmer2023teaching}, resulting in multiple key insights. 
For example, image classification models extract low-level positional and texture cues at early layers and gradually combine them into higher-level, semantic concepts at later layers~\cite{bau2017network, olah2020zoom, ghiasi2022vision}. 

However, while video transformers do share their overall architecture with image-level ViTs, the insights obtained in existing works do very little to explain their inner mechanisms. Consider, for example, the recent approach for tracking occluded objects~\cite{van2023tracking} shown in Figure~\ref{fig:Teaser} (top). To accurately reason about the trajectory of the invisible object inside the pot, texture or semantic cues alone would not suffice. What, then, are the \textit{spatiotemporal} mechanisms used by this approach? Are any of these mechanisms \textit{universal} across video models trained for different tasks? 

To answer these questions, in this work we present the Video Transformer Concept Discovery algorithm (\AlgName) -  the first concept discovery methodology for interpreting the representations of deep video transformers. We focus on concept-based interpretability~\cite{ghorbani2019towards,zhang2021invertible,fel2023craft,fel2023holistic} due to its capacity to explain the decision-making process of a complex model's distributed representations in high-level, intuitive terms. Our goal is to decompose a representation at any given layer into human-interpretable `concepts' without any labelled data (\ie concept discovery) and then rank them in terms of their importance to the model output. 

Concretely, we first group \textit{model features} at a given layer into spatiotemporal tubelets via SLIC clustering~\cite{achanta2012slic}, which serve as a basis for our analysis (Section~\ref{sec:tubelets}). Next, we cluster these tubelets across videos to discover high-level concepts~\cite{ding2008convex,lee1999learning,zhang2021invertible,fel2023craft,fel2023holistic} (Section~\ref{sec:ConceptClustering}).
The resulting concepts for an occluded object tracking method~\cite{van2023tracking} are shown in Figure~\ref{fig:Teaser} (bottom) and span a broad range of cues, including  spatiotemporal ones that detect events, like collisions, or track the containers. 

To better understand the decision-making mechanisms of video transformers, we then quantify the importance of concepts for the model's predictions.
Inspired by previous work on saliency maps~\cite{petsiuk2018rise}, we propose a novel, noise-robust approach to estimate concept importance (\Cref{sec:Importance}).
Unlike existing techniques that rely on
gradients~\cite{kim2018interpretability}, or concept occlusion~\cite{fel2023holistic}, our approach effectively handles redundancy in self-attention heads in transformer architectures.

Next, we use \AlgName to study whether there are any universal mechanisms in video transformer models, that emerge irrespective of their training objective. To this end, we extend recent work~\cite{dravid2023rosetta} to automatically identify \textit{important} concepts that are shared between several models in Section~\ref{sec:Rosetta}. We then analyze a diverse set of representations (\eg supervised, self-supervised, or video-language) and make a number of discoveries: (i) many concepts are indeed shared between models trained for different tasks; (ii) early layers tend to  form a spatiotemporal basis that underlines the rest of the information processing; (iii) later layers form object-centric video representations, even in models trained in a self-supervised way. 

We also show how \AlgName can be applied for downstream tasks. Firstly, pruning the heads of an action classification model according to their estimated importance yields a 4.3\% \textit{increase} in accuracy while reducing computation by 33\%. Secondly, object-centric concepts discovered by \AlgName can be used for video-object segmentation (VOS) and achieve strong performance on the DAVIS'16 benchmark~\cite{perazzi2016benchmark} even for self-supervised representations.

\begin{figure*}
    \centering   
\includegraphics[width=0.88\textwidth]{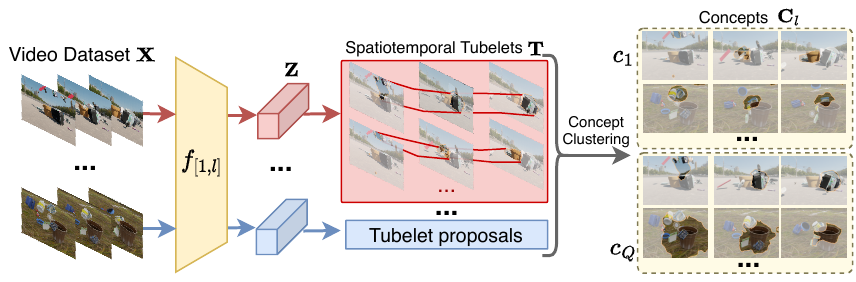}\vspace{-14pt}
    \caption{Video Transformer Concept Discovery (\AlgName) takes a dataset of videos, $\textbf{X}$, as input and passes them to a model, $f_{[1,l]}$ (shown in yellow). The set of video features, $\textbf{Z}$, are then parsed into spatiotemporal tubelet proposals, $\textbf{T}$ (shown in red), via SLIC clustering in the feature space. Finally, tubelets are clustered across the videos to discover high-level units of network representation - concepts, $\textbf{C}$ (right).}
     \vspace{-15pt}
    \label{fig:MainMethod}
\end{figure*}

\section{Related work}
Our work introduces a novel \textit{concept-based interpretability} algorithm that focuses on \textit{transformer-based representations} for \textit{video understanding}. Below, we review the most relevant works in each of these fields.

\smallsec{Concept-based interpretability} is a family of neural network interpretability methods used to understand, post-hoc, the representations that a model utilizes for a given task.
Closed-world interpretability operates under the premise of having a labeled dataset of concepts~\cite{bau2017network,kim2018interpretability}.
However, for videos, it is unclear what concepts may exist and also difficult to densely label videos even if they were known a priori. 

In contrast, unsupervised concept discovery makes no assumptions on the existence of semantic concepts and uses clustering to partition data into interpretable components within the model's feature space. ACE~\cite{ghorbani2019towards} and CRAFT~\cite{fel2023craft} segment input images into superpixels and random crops, respectively, before applying clustering at a given layer. In videos, however, the potential tubelets far outnumber image crops, prompting us to introduce a more efficient method for segmenting videos into proposals in Section~\ref{sec:tubelets}.


A necessary component of concept-based interpretability is measuring the importance (\ie fidelity) of the discovered concepts to the model. However, the aforementioned methods~\cite{kim2018interpretability,ghorbani2019towards,zhang2021invertible,fel2023craft,fel2023holistic} were developed for CNNs, and are not readily applicable to transformers. The main challenge of ranking concepts in attention heads is due to the transformers' robustness to minor perturbations in self-attention layers. To address this limitation, we introduce a new algorithm to rank the significance of any architectural unit, covering both heads and intra-head concepts in Section~\ref{sec:Importance}.

Recent work~\cite{dravid2023rosetta} identifies neurons that produce similar activation maps across various image models (including transformers). However, neurons are unable to explain the full extent of a models' distributed~\cite{elhage2022superposition} representation. In contrast, our method works on features of arbitrary dimension and is applied to video models. 

\smallsec{Interpretability of transformers}
has received significant attention recently, due to its success in a variety of computer vision tasks. Early work~\cite{raghu2021vision} contrasted vision transformer representations with CNNs (representational differences per layer, receptive fields, localization of information, etc). Other work aims to generate saliency heatmaps based on attention maps of a model~\cite{chefer2021transformer, Chefer_2021_ICCV}. Later works focused on understanding the impact of different training protocols~\cite{park2023self,walmer2023teaching} (\eg self-supervised learning (SSL) vs.\ supervised) and robustness~\cite{zhou2022understanding,qin2022understanding}. 
The features of a specific SSL vision transformer, DINO~\cite{amir2021deep}, were explored in detail and shown to have surprising utility for part-based segmentation tasks. However, none of these works address concept-based interpretability or study video representations. 

Independently, studies in natural language processing (NLP) analyzed self-attention layers~\cite{elhage2021mathematical,voita2019analyzing} and found that heads are often specialized to capture different linguistic or grammatical phenomenon. This is qualitatively seen in vision works that show dissimilar attention maps for different self-attention heads~\cite{Cordonnier2020On,Karim_2023_CVPR}. Moreover, other NLP works~\cite{michel2019sixteen,voita2019analyzing} explore the impact of removing heads and find that only a small number need to be kept to produce similar performance. Our findings agree with evidence from these works and in Section~\ref{sec:Experiments} we further demonstrate that targeted pruning of unimportant heads from video transformers can actually \textit{improve} a model's performance. 

\smallsec{Video model interpretability} is an under-explored area of research considering the recent successes of deep learning models in action recognition~\cite{kong2022human}, video object segmentation~\cite{li2013video,perazzi2016benchmark,tokmakov2023breaking,van2023tracking}, or self-supervised approaches~\cite{ranasinghe2022self,tong2022videomae,feichtenhofer2022masked,wang2023videomae,sun2023masked}. Efforts have used proxy tasks to measure the degree to which models use dynamic information~\cite{ghodrati2018video,hadji2018new,ilic2022appearance} or scene bias~\cite{li2018resound,li2019repair,choi2019can}. One method quantifies the static and dynamic information contained in a video model's intermediate representation~\cite{kowal2022deeper,kowal2022quantifying}. However, these methods can only measure one or two predefined concepts (\ie static, dynamic, or scene information) while our approach is not restricted to a subset of concepts. Another work visualizes videos that activate single neurons (or filters) in 3D CNN's via activation maximization with temporal regularization~\cite{feichtenhofer2020deep}. While this method has no restrictions on what a neuron can encode, it only applies to 3D CNNs and does not truly capture `concepts' across distributed representations (\ie feature space directions that generalize across videos).

\section{Video transformer concept discovery}


We study the problem of decomposing a video representation into a set of high-level open-world concepts and ranking their importance for the model's predictions. 
We are given a set of (RGB) videos,  $\textbf{X} \in \mathbb{R}^{N \times 3 \times T \times H \times W}$, where $N$, $T$, $H$, and $W$ denote the dataset size, time, height, and width, respectively, and an $L$ layer pretrained model, $f$. Let $f_{[r,l]}$ denote the model from layer $r$ to $l$, with $f_{[1,l]}(\textbf{X}) = \textbf{Z}_l \in \mathbb{R}^{N \times C \times T' \times H' \times W'}$ being the intermediate representation at layer $l$. 
To decompose $\textbf{Z}_l$ into a set of human-interpretable concepts, $\textbf{C}_l = \{c_1, \ldots, c_Q\}$, existing, image-level approaches~\cite{ghorbani2019towards, fel2023craft} first parse the $N$ feature maps into a set of $M$ proposals, $\textbf{T} \in \mathbb{R}^{M \times C}$ ($M > N$), where each $T_m$ corresponds to a region of the input image. These proposals are then clustered into $Q << M$ concepts in the feature space of the model to form an assignment matrix $W \in \mathbb{R}^{M \times Q}$. The importance of each concept $c_i$ to the model's prediction is quantified by a score $s_i \in [0,1]$. Performing this analysis over all layers in $f$ produces the entire set of concepts for a model, $\textbf{C} = \{\textbf{C}_1,\ldots,\textbf{C}_L\}$, together with their corresponding importance scores. 

Existing approaches are not immediately applicable to video transformers because they do not scale well and are focused on 2D CNN architectures. 
In this work, we extend concept-based interpretability to video representations. To this end, we first describe a computationally tractable proposal generation method (Section~\ref{sec:tubelets}) that operates over space-time feature volumes and outputs spatiotemporal tubelets. 
Next (Section~\ref{sec:ConceptClustering}), we adapt existing concept clustering techniques to video transformer representations. Finally, in Section~\ref{sec:Importance} we propose a novel concept importance estimation approach applicable to any architecture units, including transformer heads. 

\subsection{Concept discovery}

\subsubsection{Tubelet proposals}\label{sec:tubelets}

Previous proposal methods~\cite{ghorbani2019towards,fel2023craft,fel2023holistic} use superpixels or crops in RGB space to generate segments; however, the number of possible segments is exponentially greater for videos. Moreover, proposals in color space are unrestricted and may not align with the model's encoded information, leading to many irrelevant or noisy segments. To address these drawbacks, we instantiate proposals in \textit{feature space}, which naturally partitions a video based on the information contained within each layer (shown in Figure~\ref{fig:MainMethod}, left). 

We produce tubelets per-video via Simple Linear Iterative Clustering~\cite{achanta2012slic} (SLIC) on the spatiotemporal features as
\begin{equation}
    \textbf{T} = \text{GAP}(\textbf{B} \odot \textbf{Z}) = \text{GAP}(\text{SLIC}(\textbf{Z}) \odot \textbf{Z}) ,
\end{equation}
where $\textbf{T} \in \mathbb{R}^{M \times C} $ is the set of tubelets for the dataset, $\textbf{B} \in \{0,1\}^{C \times M \times N  \times T' \times H' \times W'}$ are spatiotemporal binary support masks obtained from SLIC, $M$ is the number of tubelets for all $N$ videos ($M>>N$), and GAP represents global average pooling over the space and time dimensions.

SLIC is an extension of the K-Means algorithm~\cite{kmeans} that controls a trade-off between cluster support regularity and adaptability, and also constrains cluster support masks to be connected. 
Together these properties produce non-disjoint tubelets that are easier to interpret for humans because they reduce the need to attend to multiple regions in a video at a time. 
Further, the pruning step in SLIC makes it more robust to the hyperparameter that controls the desired number of clusters, as it automatically prunes spurious, disconnected tubelets. 
Next, we describe our approach for grouping individual tubelets into higher-level concept clusters.

\begin{figure}
    \centering   
\includegraphics[width=0.51\textwidth]{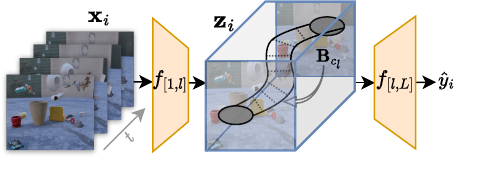}\vspace{-18pt}
    \caption{A visual representation of concept masking for a single concept. Given a video $\mathbf{x}_i$ and a concept, $c_l$, we mask the tokens of the intermediate representation $\mathbf{z}_i = f_{[1,l]}(\mathbf{x}_i)$ with the concepts' binary support masks, $\textbf{B}_{c_l}$, to obtain the perturbed prediction, $\hat{y}_i$.}
    \vspace{-13pt}
    \label{fig:CRIS_Method}
\end{figure}

\vspace{-0.13cm}
\subsubsection{Concept clustering}\label{sec:ConceptClustering}
Recent work~\cite{zhang2021invertible, fel2023craft, fel2023holistic} has used Non-Negative Matrix Factorization (NMF)~\cite{ding2008convex} to cluster proposals into concepts. Given a non-negative data matrix, $\textbf{T}^{+} \in \mathbb{R}^{M \times C}$, NMF aims to find two non-negative matrices, $\textbf{W}^{+} \in \mathbb{R}^{M \times Q}$ and $\textbf{C}^{+} \in \mathbb{R}^{Q \times C}$, such that $\textbf{T}^{+} = \textbf{W}^{+} \textbf{C}^{+}$, where $\textbf{W}^{+}$ is the cluster assignment matrix. Unfortunately, NMF cannot be applied to transformers as they use GeLU non-linearities, rather than ReLU, resulting in negative activations. 

We solve this problem by leveraging Convex Non-negative Matrix Factorization~\cite{ding2008convex} (CNMF). Despite the name, CNMF extends NMF and allows for negative input values. This is achieved by constraining the factorization such that the columns of $\textbf{W}$ are convex combinations of the columns of $\textbf{T}$, \ie each column of $\textbf{W}$ is a weighted average of the columns of $\textbf{T}$. This constraint can be written as
\begin{equation}
    \textbf{W} = \textbf{T} \textbf{G},
\end{equation}
where $\textbf{G} \in [0,1]^{C \times Q}$ and $\sum_j \textbf{G}_{i,j}=1$. To cluster a set of tubelets, $\textbf{T}$, into corresponding concepts, we optimize
\begin{equation}
    (\textbf{G}^*,\textbf{C}^*) = \argmin_{\textbf{C}>0,\textbf{G}>0} || \textbf{T} - \textbf{T} \textbf{G} \textbf{C} ||^2,
\end{equation}
where the final set of concepts are the rows of the matrix $\textbf{C}$, \ie concept centroid $c_i$ is the $i^{th}$ row of $\textbf{C}$ (Figure~\ref{fig:MainMethod}, right).

\subsection{Concept importance}\label{sec:Importance}
Given a set of discovered concepts, we now aim to quantify their impact on model performance.
One approach, shown in Figure~\ref{fig:CRIS_Method}, is to mask out each concept independently and rank the importance based on the drop in performance~\cite{fel2023holistic}. Formally, let $c_l$ be a single target concept, and $\textbf{B}_{c_l} \in \{0,1\}^{C \times M \times N  \times T' \times H' \times W'}$ the corresponding binary support masks over $\mathbf{X}$. It can then be masked in layer $l$ via
\begin{equation}\label{eq:ConceptMaskSingle}
    \hat y = f_{[l,L]}( \textbf{Z}_l \odot (1-\textbf{B}_{c_l})).
\end{equation}

While this approach works well for CNNs~\cite{fel2023holistic}, transformers are robust to small perturbations within self-attention layers~\cite{michel2019sixteen,voita2019analyzing}. Therefore, single concept masking has little effect on performance (shown by results in Figure~\ref{fig:AttributionValidation}). Instead, we mask \textit{a high percentage} of sampled concepts in parallel (across all layers and heads) and then empirically validate in Section~\ref{sec:ExpValidation} that averaging the results over thousands of samples produces valid concept rankings.

Formally, we propose \textbf{C}oncept \textbf{R}andomized \textbf{I}mportance \textbf{S}ampling (\ConceptRankAlgName), a robust method to compute importance for any unit of interest. 
To this end, we first randomly sample $K$ different concept sets, such that each $\textbf{C}^k \subset \textbf{C}$. We then define $\textbf{C}^k_l$ as the set of concepts in $\textbf{C}^{k}$ discovered at layer $l$, with $\textbf{B}_{\textbf{C}^k_l}$ denoting the corresponding binary support masks. We mask out every concept at every layer via
\begin{equation}\label{eq:ConceptMaskMultiple2}
    \hat y_k = g(\tilde{\textbf{B}}_{\textbf{C}^k_L} \odot f_{[L-1,L]} (\cdots( \tilde{\textbf{B}}_{\textbf{C}^k_{1}} \odot f_{[0,1]} (\mathbf{X}) ) ) ),
\end{equation}
where $g(\cdot)$ is the prediction head (\eg an MLP) and $\tilde{\textbf{B}}$ denotes the inverse mask (\ie $1-\textbf{B}$). Finally, we calculate the importance of each concept, $c_i$,  via
\begin{equation}
    s_i = \frac{1}{K} \sum_k^K (\mathbb{D}(\tilde y, y)-\mathbb{D}(\hat y_k, y)) \mathbbm{1}_{c_i \in \textbf{C}^k},
\end{equation}
where $\tilde y$ is the original prediction without any masking and $\mathbb{D}$ is a metric quantifying performance (\eg accuracy).

\section{Understanding transformers with VTCD}\label{sec:UnderstandingTransformers}

Our algorithm facilitates the identification of concepts within any unit of a model and quantifying their significance in the final predictions; however, this is insufficient to fully represent the computations performed by a video transformer. It is also crucial to understand how these concepts are employed in the model's information flow.

As several recent works have shown~\cite{elhage2021mathematical,olsson2022context}, the residual stream of a transformer serves as the backbone of the information flow. Each self-attention block then reads information from the residual stream with a linear projection, performs self-attention operations to process it, and finally writes the results back into the residual stream. Crucially, self-attention processing is performed individually for each head with several studies showing, both in vision~\cite{amir2021deep,dosovitskiy2020image,Karim_2023_CVPR} and NLP~\cite{voita2019analyzing}, that different self-attention heads capture distinct information. In other word, heads form the basis of the transformer representation.


A closer analysis of the concepts found in the heads of the TCOW model~\cite{van2023tracking} with \AlgName allows us to identify several information processing patterns in that model. In particular, Figure~\ref{fig:Teaser} shows that the heads in early layers group input tokens based on their spatiotemporal positions. This information is then used to track objects and identify events in mid-layers, and later layers utilize mid-layer representations to reason about occlusions. Next, we study which of these mechanisms are \textit{universal} across video transformers trained on different datasets and objectives (\textit{cf.}~\cite{dravid2023rosetta}).

\subsection{Rosetta concepts}\label{sec:Rosetta}
Inspired by previous work~\cite{dravid2023rosetta}, we mine for \textit{Rosetta concepts} that are shared between models and represent the same information. 
The key to identifying Rosetta units is a robust metric, $R$, where a higher $R$-score corresponds to the two units having a larger amount of shared information. This previous work~\cite{dravid2023rosetta} focused on finding such neurons in image models based on correlating their activation maps. We instead measure the similarity between concepts (\ie distributed representations) via the mean Intersection over Union (mIoU) of the concepts' support.

Formally, we mine Rosetta concepts by first applying \AlgName to a set of $D$ models $\{f^1,\ldots,f^D\}$, resulting in discovered concepts, $\textbf{C}^j = \{c^j_1, \ldots, c^j_i\}$, and importance scores, $\textbf{S}^j = \{s^j_1, \ldots, s^j_i\}$, for each model $f^j$. We then aim to measure the similarity between all concept $D$-tuples from the models. Given a set of $D$ concepts, $\{c^1_i,\ldots,c^D_i\}$, and corresponding binary support masks, $\{\textbf{B}^1_i,\ldots,\textbf{B}^D_i\}$, we define the similarity score of these concepts as
\begin{equation}
    R^D_i  = \frac{|\textbf{B}^{1}_i \cap \cdots \cap \textbf{B}^{D}_i|}{|\textbf{B}^{1}_i \cup \cdots \cup \textbf{B}^{D}_i|}.
    \label{eq:r}
\end{equation}

Naively computing the similarity between all $D$-tuples results in an exponential number of computations and is intractable for even small $D$. To mitigate these issues, we exclude two types of concepts: (i) unimportant ones and (ii) those with a low $R$-score among $d$-tuples, where $d < D$. More specifically, we only consider the most important $\epsilon \%$ of concepts from each model. We then iterate over $d \in \{2,\ldots,D\}$ and filter out concepts that have $R$-scores less than $\delta$ for all d-tuples in which it participates. Formally, the filtered Rosetta d-concept scores are defined as
\begin{equation}
  \textbf{R}^d_{\epsilon,\delta} = \{ R^d_i \hspace{0.05cm} | \hspace{0.05cm} R^d_i > \delta \hspace{0.05cm} \forall R^d_i \in \textbf{R}^d_{\epsilon} \},
\label{eq:r2}
\end{equation}
where $\textbf{R}^d_\epsilon$ is the set of all $R$-scores among $d$ concepts after the $\epsilon$ importance filtering. 
This results in a significantly smaller pool of candidates for the next stage, $d+1$, reducing the overall computational complexity of the algorithm. Finally, as some concepts may reside in a subset of the models but are still interesting to study, we examine the union of all important and confident Rosetta d-concepts corresponding to $R$-scores $\textbf{R}^2_{\epsilon,\delta} \cup \cdot \cdot \cdot \cup\textbf{R}^D_{\epsilon,\delta}$.

\section{Experiments}\label{sec:Experiments}
We evaluate our concept discovery algorithm quantitatively and qualitatively across a variety of models and tasks.

\smallsec{Datasets.}
We primarily use two datasets in our experiments: TCOW Kubric~\cite{van2023tracking} and Something-Something-v2 (SSv2)~\cite{goyal2017something}.
The former is a synthetic, photo-realistic dataset of 4,000 videos with randomized object location and motion, based on the Kubric synthetic video generator~\cite{greff2022kubric}.
This dataset is intended for semi-supervised VOS (semi-VOS) through occlusions. SSv2 contains 220,847 real videos intended for fine-grained action recognition.
Each sample is a crowdsourced video of a person-object interaction (\ie doing something to something). Unlike other video classification benchmarks~\cite{carreira2017quo,soomro2012ucf101}, temporal reasoning is fundamental for distinguishing SSv2 actions, making it an ideal choice for analyzing spatiotemporal mechanisms.

\smallsec{Models.}
We evaluate four models with public pretrained checkpoints:
(i) TCOW~\cite{van2023tracking} trained on Kubric for semi-VOS,
(ii) VideoMAE~\cite{tong2022videomae} trained on SSv2 for action classification (Supervised VideoMAE),
(iii) VideoMAE self-supervised on SSv2 (SSL VideoMAE),
and
(iv) InternVideo~\cite{wang2022internvideo}, a video-text model trained contrastively on 12M clips from eight video datasets and 100M image-text pairs from LAION-400M~\cite{schuhmann2021laion}.
As TCOW requires a segmentation as input, when applying it to SSv2, we manually label the most salient object in the initial frame.
We focus our analysis on the first two models, and use the last two to validate the universality of our Rosetta concepts.

\smallsec{Implementation details.}
For all experiments, we run VTCD on 30 randomly sampled videos and discover concepts from all heads and layers.
We focus on keys in the self-attention heads, as they produce the most meaningful clusters per prior work~\cite{amir2021deep}. We present results with queries and values on the project page, with further implementation details.

\smallsec{\AlgName target metrics.}
To discover and rank concepts, \AlgName requires a target evaluation metric.
For TCOW Kubric, we use the intersection-over-union (IoU) between the predicted and the groundtruth masks.
For SSv2, we use classification accuracy for a target class.


\begin{table}[t]
    \centering
    \resizebox{0.48\textwidth}{!}{
    \begin{tabular}{lcccc}
        \toprule
         & \multicolumn{2}{c}{TCOW} & \multicolumn{2}{c}{VideoMAE} \\
         Model & Positive $\downarrow$ & Negative $\uparrow$ & Positive $\downarrow$ & Negative $\uparrow$ \\
         \midrule
         Baseline + Occ & 0.174 & 0.274 & 0.240 & 0.300 \\
         Baseline + CRIS & 0.166&0.284&0.157&0.607 \\
         \AlgName (Ours) & \textbf{0.102} & \textbf{0.288} & \textbf{0.094} & \textbf{0.625} \\
         \bottomrule
    \end{tabular}
    }\vspace{-0.33cm}
    \caption{Comparison of our tubelet proposal approach to the widely used random crop tubelets~\cite{yeh2020completeness,fel2023craft,fel2023holistic} with occlusion-based importance~\cite{fel2023holistic} (Baseline + Occ) for both TCOW~\cite{van2023tracking} and VideoMAE~\cite{tong2022videomae}. Our tubelets result in concepts that are more faithful to the model's representations even when the baseline is equipped with our concept scoring algorithm (Baseline + CRIS).}
    \vspace{-0.4cm}
    \label{tab:TubeletValidation}
\end{table}

\subsection{Quantitative evaluation}\label{sec:ExpValidation}
This section presents a quantitative validation of \AlgName's key components compared to various baselines. We first assess our tubelet proposal methodology, followed by a comparison of CRIS with other concept importance methods. We follow the standard evaluation protocol, and measure the \textit{fidelity} of the discovered concepts~\cite{ghorbani2019towards,zhang2021invertible,fel2023craft}. To this end, we calculate attribution curves~\cite{ghorbani2019towards,fel2023craft,fel2023holistic}, where concepts are removed from a model in either most-to-least order (\textit{positive} perturbation), or least-to-most (\textit{negative} perturbation). The intuition is that concepts, and corresponding importance scores, with higher fidelity will have a steeper performance decrease when removing the most important concepts, and vice-versa for the reverse order.

\smallsec{Tubelet validation. }
Unlike previous methods that partition the inputs into proposals in the pixel space, \AlgName generates the proposals via SLIC~\cite{achanta2012slic} clustering in the model's feature space. We ablate this design choice by comparing \AlgName with a strong  \textit{random cropping} baseline used by recent image-based concept discovery methods (ConceptSHAP~\cite{yeh2020completeness}, CRAFT~\cite{fel2023craft}, Lens~\cite{fel2023holistic}), shown as `Baseline + Occ' in Table~\ref{tab:TubeletValidation}.
In all cases, \AlgName yields concepts that are more faithful to the model's representation. To further isolate the effect of proposals on the performance, we equip the baseline with our concept importance estimation approach (shown as `Baseline + CRIS'). We observe that \AlgName still outperforms this strong baseline in all settings, validating that generating tubelet proposals in the feature space of the transformer indeed yields concepts that are more faithful to the model's representation.

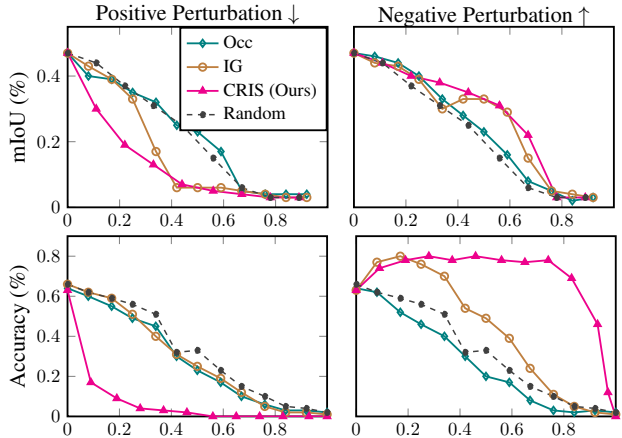
\begin{figure}[t]
	\begin{center}
     \centering 
\resizebox{0.48\textwidth}{!}{
\hspace{0.1cm}
\begin{tikzpicture} 
                 \begin{axis}[
                 line width=1.0,
                 title={Positive Perturbation $\downarrow$},
                 title style={at={(axis description cs:0.5,1.1)},anchor=north,font=\large},
                 ylabel={mIoU (\%)},
                 xmin=0, xmax=1,
                 ymin=0, ymax=0.55,
                 xtick={0,0.2,0.4,0.6,0.8},
                 x tick label style={font=\normalsize, rotate=0, anchor=north},
                 y tick label style={font=\normalsize},
                 y label style={at={(axis description cs:0,.5)},anchor=south,font=\at={(axis description cs:0.5,0.09)},anchor=north,font=\large},
                 width=6.5cm,
                 height=5cm, 
                 ymajorgrids=false,
                 xmajorgrids=false,
                 major grid style={dotted,teal!20!black},
                 legend columns=1,
                 legend style={
                  nodes={scale=1.1, transform shape},
                  cells={anchor=west},
                  legend style={at={(1,1)},anchor=north east,row sep=0.01pt}, font =\small}
             ]
            \addlegendentry{Occ}
            \addplot[line width=1pt, mark size=2pt, color=teal, mark=diamond,error bars/.cd, y dir=both, y explicit,]
                     coordinates {
(0.00,	0.47)
(0.08,	0.40)
(0.17,	0.39)
(0.25,	0.35)
(0.34,	0.32)
(0.42,	0.25)
(0.50,	0.23)
(0.59,	0.17)
(0.67,	0.05)
(0.76,	0.04)
(0.84,	0.04)
(0.92,	0.04)
};  

            \addlegendentry{IG}
            \addplot[line width=1pt, mark size=2pt, color=brown, mark=o,error bars/.cd, y dir=both, y explicit,]
                     coordinates {
(0.00,0.47)
(0.08,0.43)
(0.17,0.39)
(0.25,0.33)
(0.34,0.17)
(0.42,0.06)
(0.50,0.06)
(0.59,0.06)
(0.67,0.05)
(0.76,0.04)
(0.84,0.03)
(0.92,0.03)
};   

            \addlegendentry{CRIS (Ours)}
            \addplot[line width=1pt, mark size=2pt, color=magenta, mark=triangle*,error bars/.cd, y dir=both, y explicit,]
                     coordinates {
(0.00,0.47)
(0.11,0.30)
(0.22,0.19)
(0.33,0.13)
(0.44,0.07)
(0.56,0.05)
(0.67,0.04)
(0.78,0.03)
(0.89,0.03)
};  
            \addlegendentry{Random}
          \addplot[dashed,line width=0.8pt, mark size=1.5pt, color=darkgray, mark=*,error bars/.cd, y dir=both, y explicit,]
                     coordinates {
(0.00,0.47)
(0.11,0.44)
(0.22,0.37)
(0.33,0.31)
(0.44,0.25)
(0.56,0.15)
(0.67,0.06)
(0.78,0.03)
(0.89,0.03)
};   

              \end{axis}
\end{tikzpicture}
\hspace{0.1cm}
\begin{tikzpicture} 
                 \begin{axis}[
                 line width=1.0,
                 title={Negative Perturbation $\uparrow$},
                 title style={at={(axis description cs:0.5,1.09)},anchor=north,font=\large},
                 xmin=0, xmax=1,
                 ymin=0, ymax=0.55,
                 xtick={0,0.2,0.4,0.6,0.8},
                 ymajorticks=false,
                 x tick label style={font=\normalsize, rotate=0, anchor=north},
                 y tick label style={font=\normalsize},
                 x label style={at={(axis description cs:0.5,0.09)},anchor=north,font=\normalsize},
                 y label style={at={(axis description cs:0.2,.5)},anchor=south,font=\normalsize},
                 width=6.5cm,
                 height=5cm, 
                 ymajorgrids=false,
                 xmajorgrids=false,
                 major grid style={dotted,darkteal!20!black},
             ]
            \addplot[line width=1pt, mark size=2pt, color=teal, mark=diamond,error bars/.cd, y dir=both, y explicit,]
                     coordinates {
(0.00,	0.47)
(0.08,	0.46)
(0.17,	0.44)
(0.25,	0.40)
(0.34,	0.33)
(0.42,	0.28)
(0.50,	0.23)
(0.59,	0.16)
(0.67,	0.08)
(0.76,	0.05)
(0.84,	0.02)
(0.92,	0.03)
};  

            \addplot[line width=1pt, mark size=2pt, color=brown, mark=o,error bars/.cd, y dir=both, y explicit,]
                     coordinates {
(0.00,	0.47)
(0.08,	0.44)
(0.17,	0.43)
(0.25,	0.39)
(0.34,	0.30)
(0.42,	0.33)
(0.50,	0.33)
(0.59,	0.29)
(0.67,	0.15)
(0.76,	0.05)
(0.84,	0.04)
(0.92,	0.03)
};   

            \addplot[line width=1pt, mark size=2pt, color=magenta, mark=triangle*,error bars/.cd, y dir=both, y explicit,]
                     coordinates {
(0.00,0.47)
(0.11,0.44)
(0.22,0.40)
(0.33,0.38)
(0.44,0.35)
(0.56,0.31)
(0.67,0.22)
(0.78,0.03)
(0.89,0.03)
};   

            \addplot[dashed,line width=0.8pt, mark size=1.5pt, color=darkgray, mark=*,error bars/.cd, y dir=both, y explicit,]
                     coordinates {
(0.00,0.47)
(0.11,0.44)
(0.22,0.37)
(0.33,0.31)
(0.44,0.25)
(0.56,0.15)
(0.67,0.06)
(0.78,0.03)
(0.89,0.03)
};   
              \end{axis}
\end{tikzpicture}
}
\resizebox{0.48\textwidth}{!}{
\hspace{0.1cm}
\begin{tikzpicture} 
                 \begin{axis}[
                 line width=1.0,
                 title style={at={(axis description cs:0.5,1.09)},anchor=north,font=\large},
                 ylabel={Accuracy (\%)},
                 xmin=0, xmax=1,
                 ymin=0, ymax=0.9,
                 xtick={0,0.2,0.4,0.6,0.8},
                 x tick label style={font=\normalsize, rotate=0, anchor=north},
                 y tick label style={font=\normalsize},
                 y label style={at={(axis description cs:0,.5)},anchor=south,font=\at={(axis description cs:0.5,0.09)},anchor=north,font=\large},
                 width=6.5cm,
                 height=5cm, 
                 ymajorgrids=false,
                 xmajorgrids=false,
                 major grid style={dotted,teal!20!black},
                 legend columns=1,
                 legend style={
                  nodes={scale=1.2, transform shape},
                  cells={anchor=west},
                  legend style={at={(1,1)},anchor=north east,row sep=0.01pt}, font =\small}
             ]
            \addplot[line width=1pt, mark size=2pt, color=teal, mark=diamond]
                     coordinates {
(0.00,	0.64)
(0.08,	0.60)
(0.17,	0.55)
(0.25,	0.49)
(0.34,	0.45)
(0.42,	0.30)
(0.50,	0.23)
(0.59,	0.17)
(0.67,	0.10)
(0.76,	0.06)
(0.84,	0.03)
(0.92,	0.03)
(1.00,	0.02)
};

            \addplot[line width=1pt, mark size=2pt, color=brown, mark=o,error bars/.cd, y dir=both, y explicit,]
                     coordinates {
(0.00,	0.66)
(0.08,	0.62)
(0.17,	0.59)
(0.25,	0.51)
(0.34,	0.40)
(0.42,	0.31)
(0.50,	0.25)
(0.59,	0.19)
(0.67,	0.12)
(0.76,	0.05)
(0.84,	0.02)
(0.92,	0.02)
(1.00,	0.01)
};  
            \addplot[line width=1pt, mark size=2pt, color=magenta, mark=triangle*,error bars/.cd, y dir=both, y explicit,]
                     coordinates {
(0.00,0.63)
(0.09,0.17)
(0.19,0.09)
(0.28,0.04)
(0.37,0.03)
(0.46,0.02)
(0.56,0.00)
(0.65,0.00)
(0.74,0.00)
(0.83,0.00)
(0.93,0.00)
(1.00,0.00)
};  
            \addplot[dashed,line width=0.8pt, mark size=1.5pt, color=darkgray, mark=*,error bars/.cd, y dir=both, y explicit,]
                     coordinates {
(0.00,	0.66)
(0.08,	0.62)
(0.17,	0.59)
(0.25,	0.56)
(0.34,	0.51)
(0.42,	0.32)
(0.50,	0.33)
(0.59,	0.23)
(0.67,	0.15)
(0.76,	0.10)
(0.84,	0.05)
(0.92,	0.04)
(1.00,	0.02)
};   

              \end{axis}
\end{tikzpicture}
\hspace{0.1cm}
\begin{tikzpicture} 
                 \begin{axis}[
                 line width=1.0,
                 title style={at={(axis description cs:0.5,1.1)},anchor=north,font=\Large},
                 xmin=0, xmax=1,
                 ymin=0, ymax=0.9,
                 xtick={0,0.2,0.4,0.6,0.8},
                 ymajorticks=false,
                 x tick label style={font=\normalsize, rotate=0, anchor=north},
                 y tick label style={font=\normalsize},
                 x label style={at={(axis description cs:0.5,0.09)},anchor=north,font=\normalsize},
                 y label style={at={(axis description cs:0.2,.5)},anchor=south,font=\normalsize},
                 width=6.5cm,
                 height=5cm, 
                 ymajorgrids=false,
                 xmajorgrids=false,
                 major grid style={dotted,teal!20!black},
             ]

            \addplot[line width=1pt, mark size=2pt, color=teal, mark=diamond,error bars/.cd, y dir=both, y explicit,]
                     coordinates {
(0.00,	0.64)
(0.08,	0.62)
(0.17,	0.52)
(0.25,	0.46)
(0.34,	0.40)
(0.42,	0.30)
(0.50,	0.20)
(0.59,	0.17)
(0.67,	0.08)
(0.76,	0.03)
(0.84,	0.02)
(0.92,	0.03)
(1.00,	0.02)
};   
            \addplot[line width=1pt, mark size=2pt, color=brown, mark=o,error bars/.cd, y dir=both, y explicit,]
                     coordinates {
(0.00,	0.63)
(0.08,	0.77)
(0.17,	0.80)
(0.25,	0.76)
(0.34,	0.70)
(0.42,	0.54)
(0.50,	0.49)
(0.59,	0.39)
(0.67,	0.24)
(0.76,	0.11)
(0.84,	0.05)
(0.92,	0.02)
(1.00,	0.01)
};   

            \addplot[line width=1pt, mark size=2pt, color=magenta, mark=triangle*,error bars/.cd, y dir=both, y explicit,]
                     coordinates {
(0.00,0.63)
(0.09,0.74)
(0.19,0.78)
(0.28,0.80)
(0.37,0.78)
(0.46,0.80)
(0.56,0.78)
(0.65,0.77)
(0.74,0.78)
(0.83,0.69)
(0.93,0.46)
(0.97,0.12)
(1.00,0.00)
};  

            \addplot[dashed,line width=0.8pt, mark size=1.5pt, color=darkgray, mark=*,error bars/.cd, y dir=both, y explicit,]
                     coordinates {
(0.00,	0.66)
(0.08,	0.62)
(0.17,	0.59)
(0.25,	0.56)
(0.34,	0.51)
(0.42,	0.32)
(0.50,	0.33)
(0.59,	0.23)
(0.67,	0.15)
(0.76,	0.10)
(0.84,	0.05)
(0.92,	0.04)
(1.00,	0.02)
};   
              \end{axis}
\end{tikzpicture}}
\end{center}
\vspace{-0.6cm}
\caption{Attribution curves for every layer of TCOW trained on Kubric (top) and VideoMAE trained on SSv2 (bottom). We remove concepts from most-to-least (left) or least-to-most important (right). CRIS produces better concept importance than methods based on single concept occlusions (Occ) or gradients (IG).
}\label{fig:AttributionValidation}
\vspace{-10pt}
\end{figure}

\begin{table}[t]
    \centering
    \begin{tabular}{lll}
        \toprule
         Model & Accuracy $\uparrow$ & GFLOPs $\downarrow$ \\
         \midrule
         Baseline & 37.1 & 180.5 \\
         \midrule 
         VTCD 33\% Pruned & \textbf{41.4} & 121.5 \\
         VTCD 50\% Pruned & 37.8 & \textbf{91.1} \\
         \bottomrule
    \end{tabular}
    \vspace{-0.25cm}
    \caption{Pruning unimportant heads in VideoMAE results in improved efficiency and accuracy when targeting a subset of classes. Here, we target the six SSv2 classes containing types of spills.}
    \vspace{-0.5cm}
    \label{tab:ApplicationResults}
\end{table}

\begin{figure*}
    \centering   
\includegraphics[width=\textwidth]{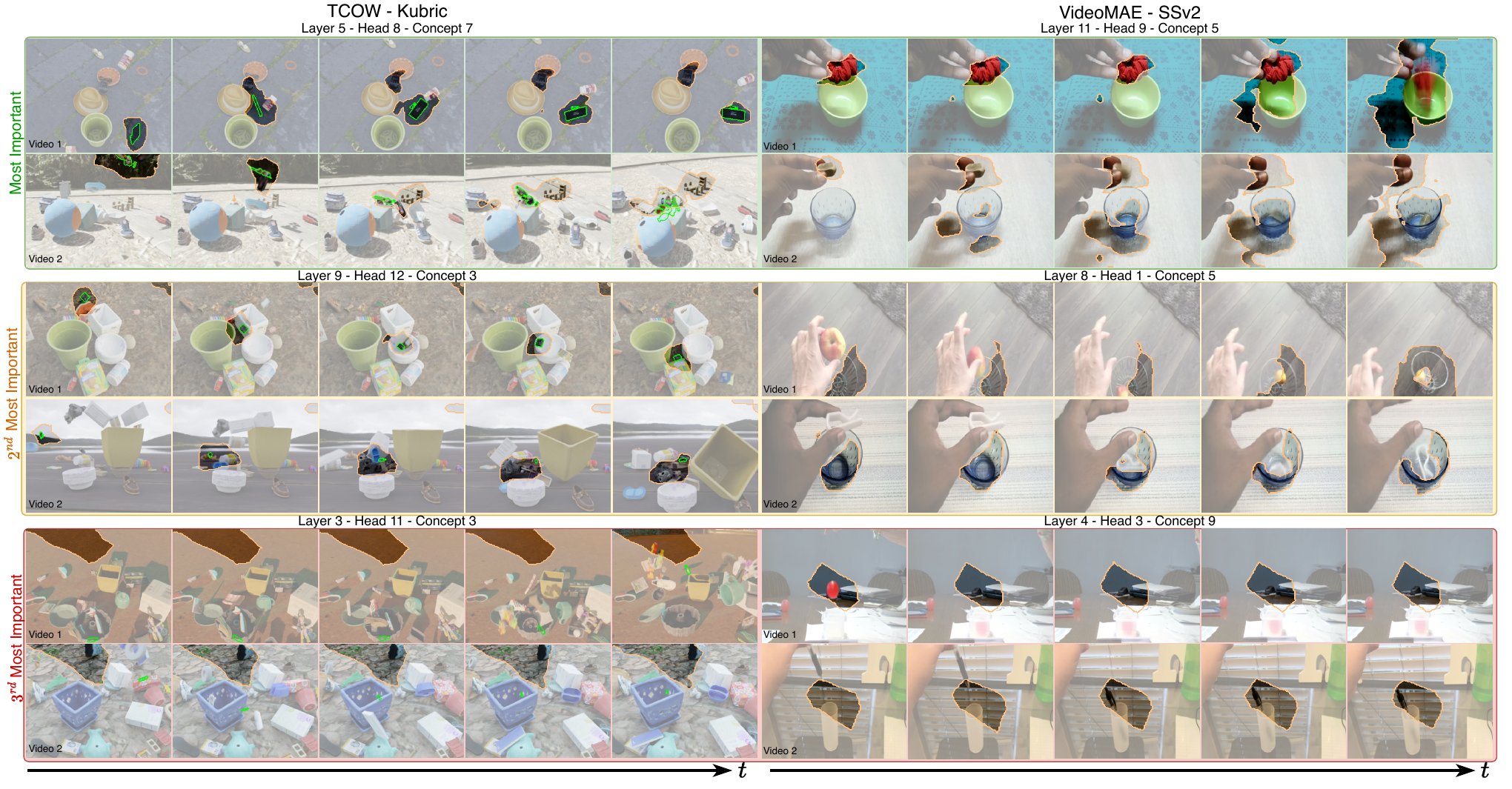}\vspace{-0.4cm}
    \caption{The top-3 most important concepts for the TCOW model trained on Kubric (left) and VideoMAE trained on SSv2 for the target class \textit{dropping something into something} (right). Two videos are shown for each concept and the query object is denoted with a green border in Kubric. For TCOW, the $1^{st}$ and $2^{nd}$ (top-left, middle-left) most important concepts track multiple objects including the target and the distractors. For VideoMAE, the top concept (top-right) captures the object and dropping event (\ie hand, object and container) while the $2^{nd}$ most important concept (middle-right) captures solely the container. Interestingly, for both models and tasks, the third most important concept (bottom) is a temporally invariant tubelet. See Section~\ref{sec:RosettaExp} for further discussion (and the \href{https://youtu.be/AsvTkcdvdC4}{video} for full results).} 
    \vspace{-0.3cm}
    \label{fig:ConceptExamples1}
\end{figure*}

\begin{figure}
    \centering   
\includegraphics[width=0.45\textwidth]{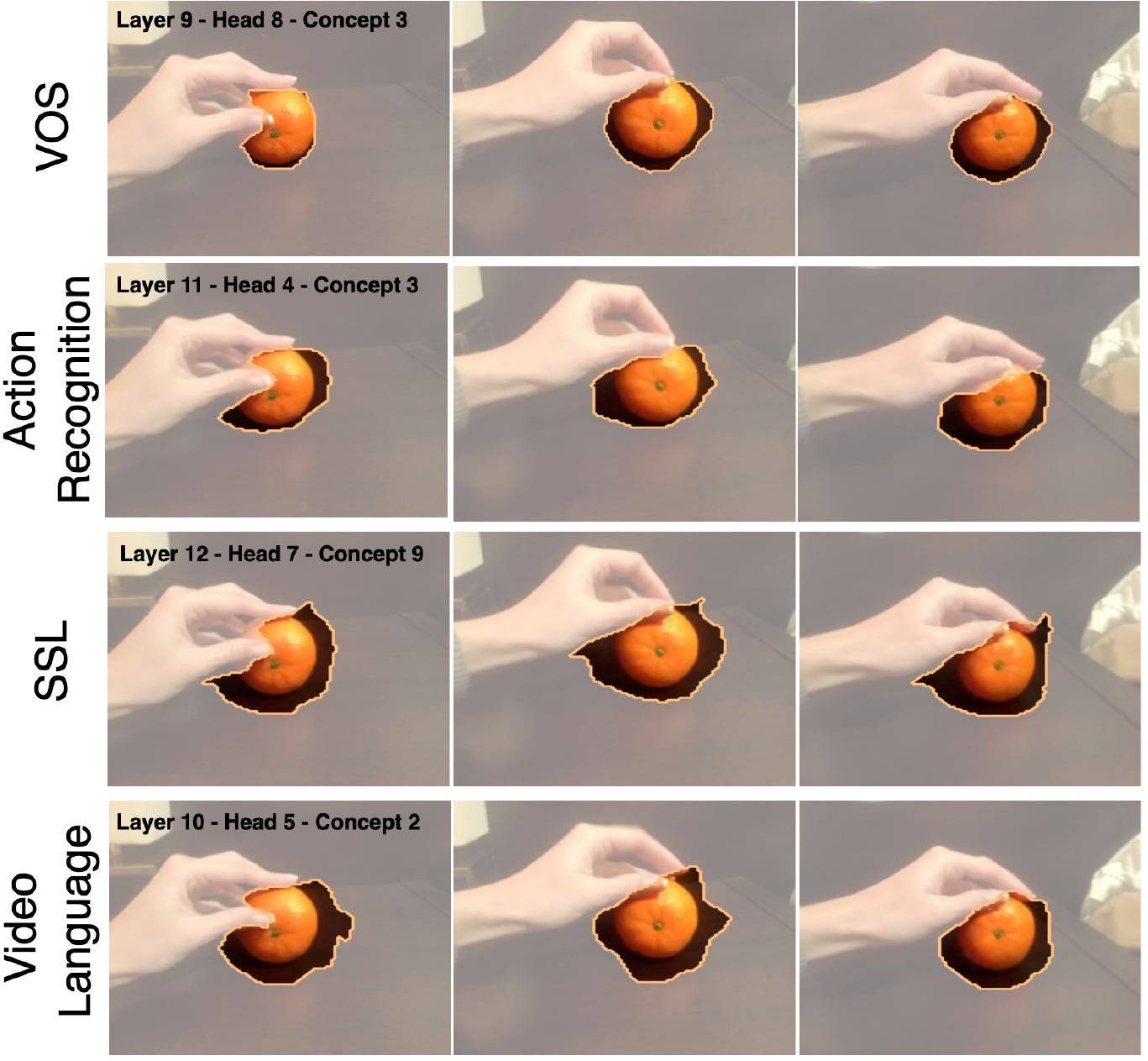}
\vspace{-0.2cm}
    \caption{A sample Rosetta concept found in four models trained for different tasks. Interestingly, we find object-centric representations in all the models 
    (see \href{https://youtu.be/Nt-e41ojAHI}{video} for full results).
    }
    \vspace{-0.6cm}
    \label{fig:RosettaValidate}
\end{figure}


\smallsec{Concept important evaluation. }
In Figure~\ref{fig:AttributionValidation}, we plot concept attribution curves of our method for TCOW for Kubric (top) and supervised VideoMAE for SSv2, targeting 10 randomly sampled classes and averaging results (bottom). In addition, we report several baselines: (i) concept removal in a random order, (ii) standard, occlusion-based concept importance estimation~\cite{fel2023holistic}, and (iii) a gradient based approach~\cite{kim2018interpretability,fel2023holistic}.
In all cases, CRIS produces a more viable importance ranking, dramatically outperforming both random ordering and the occlusion baseline. 
The integrated gradients method performs similarly to ours for TCOW, but is much worse for the action classification VideoMAE.

\begin{figure*}[t]
    \centering   
\includegraphics[width=\textwidth]{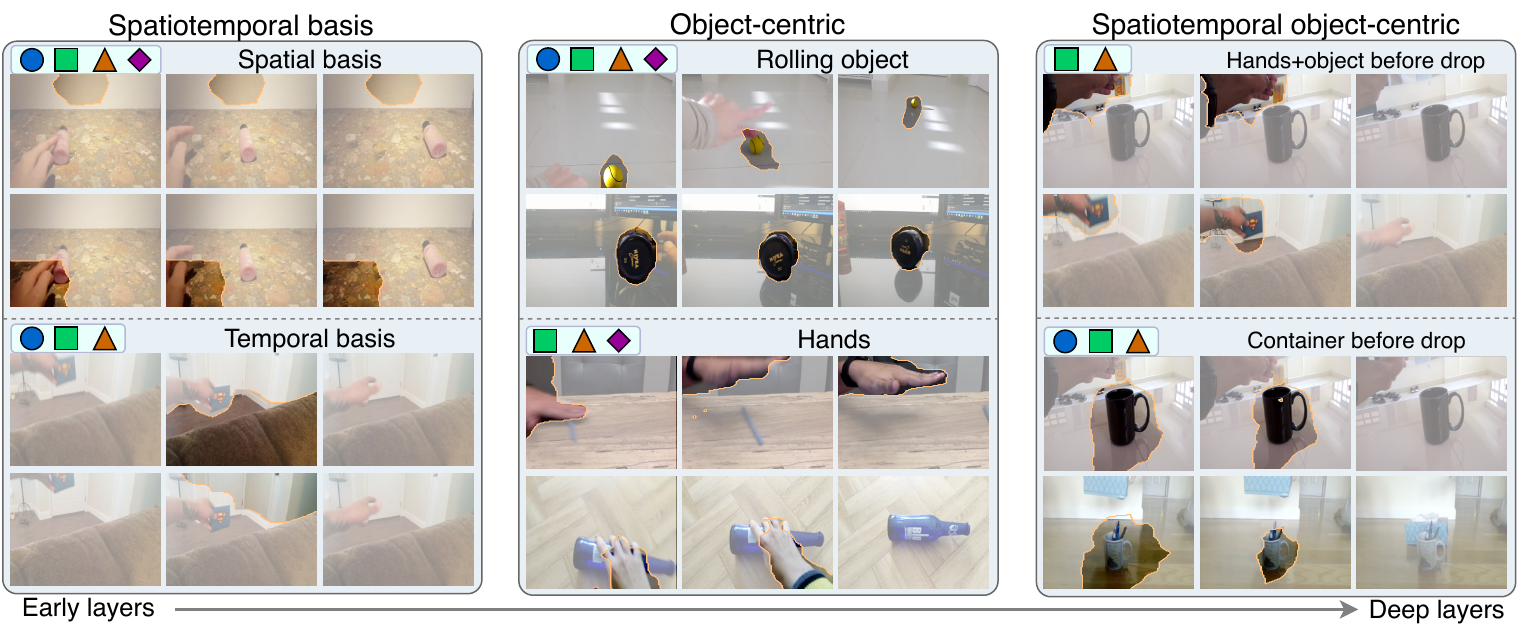}\vspace{-0.47cm}
    \caption{Universal concepts emerge in video transformers despite being trained for different tasks. Early layers encode spatiotemporal positional information. Middle layers track various objects. Deep layers capture fine-grained spatiotemporal concepts, \eg~related to occlusion reasoning (see \href{https://youtu.be/pvH5QrwO4Ro}{video} for full results). 
    Legend: \tikzbluecircle \hspace{0.05cm} TCOW,  \tikzgreensquare \hspace{0.05cm} Supervised VideoMAE, \tikzorangetriangle \hspace{0.05cm} SSL VideoMAE, \tikzpurplediamond \hspace{0.05cm} InternVideo.}
    \vspace{-0.4cm}
    \label{fig:RosettaConcepts}
\end{figure*}

Notably, we observe that the performance actually \textit{increases} for VideoMAE when up to 70\% of the least important concepts are removed. Recall that SSv2 is a fine-grained action classification dataset. \AlgName removes concepts that are irrelevant for the given class, hence increasing the robustness of the model's predictions. We further quantify this effect in Table~\ref{tab:ApplicationResults}, where we focus on the six classes where a `spill' happens (listed in the appendix). We then use CRIS to rank the \textit{heads} in VideoMAE according to their effect on performance using the training set and report the results for pruning the least important ones on the validation set. Pruning 33\% of the heads actually \textit{improves} the accuracy by 4.3\% while reducing FLOPS from 180.5 to 121.5. Further removing 50\% of the heads retains the original performance (+0.7\%) and reduces FLOPs to 91.1. These results show that one can tune the trade-off between performance and computation using \AlgName for pruning.

\subsection{Qualitative analysis}\label{sec:GeneralConcepts}
We have seen that the importance assigned to concepts discovered by VTCD aligns with the accuracy of the model.
We now assess the concepts qualitatively.
To this end, Figure~\ref{fig:ConceptExamples1} shows two representative videos for the top three most important concepts for the TCOW and VideoMAE models for the class \textit{dropping something into something}.

For TCOW, the most important concept occurs in layer five and tracks the target object. Interestingly, the same concept highlights objects with similar appearance and 2D position to the target.
This suggests that the model solves the disambiguation problem by first identifying possible distractors in mid-layers (\ie five) and then using this information to more accurately track the target in final layers.
In fact, the second most important concept, occurring in layer nine, tracks the target object throughout the video. 

For VideoMAE, the most important concept highlights the object until it is dropped, after which the object and its container are highlighted.
The second concept clearly captures the container, but notably not the object itself, making a ring-like shape.
These concepts identify an important mechanism for differentiating similar classes (\eg \textit{dropping something into/behind/in-front of something}). 

The third concept for each model captures similar information, occurring in early layers: a temporally invariant, spatial support. This corroborates research~\cite{amir2021deep,ghiasi2022vision} suggesting that positional information processing occurs early in the model, acting as a reference frame between semantic information and the tokens themselves. 
We now turn to Rosetta concepts: concepts \textit{shared} across multiple models.

\vspace{-0.25em}
\subsection{Rosetta concepts}\label{sec:RosettaExp}

We begin by applying VTCD to the four models, 
targeting two classes chosen due to their dynamic nature: \textit{rolling something across a flat surface} and \textit{dropping something behind something}.
We then mine Rosetta concepts following Section~\ref{sec:Rosetta}, with $\delta=0.15$ and $\epsilon=15\%$ in all experiments. 
The resulting Rosetta 4-concepts contains 40 tuples with an average $R$-score (Equation~\ref{eq:r}) of 17.1. Note the average $R$-score between all possible 4-concepts is 0.6, indicating the significance of the selected matches.
Figure~\ref{fig:RosettaValidate} visualizes one mined Rosetta 4-concept, which captures an object tracking representation in the late layers. This demonstrates that universal representations indeed exist between all four models. 
Our project website shows more shared concepts as videos.

Next, we qualitatively analyze all Rosetta d-concept with $d \in \{2, 3, 4\}$ at various layers. Figure~\ref{fig:RosettaConcepts} shows a representative sample.
In early layers, we find that the models learn spatiotemporal basis representations (Figure~\ref{fig:RosettaConcepts}, left): they decompose the video space-time volume into connected regions, facilitating higher-level reasoning in later layers. This is consistent with prior works that show spatial position is encoded in early layers of image transformers~\cite{amir2021deep,ghiasi2022vision}.


In the mid-layers (Figure~\ref{fig:RosettaConcepts}, middle), we find that, among other things, all the models learn to localize and track individual objects. This result introduces a new angle to the recently developed field of object-centric representation learning~\cite{locatello2020object,bao2022discovering,seitzer2022bridging,gao2017object}: it invites us to explore how specialized approaches contribute, given that object concepts naturally emerge in video transformers. 
In addition, all models, except for synthetically trained TCOW, develop hand tracking concepts, confirming the importance of hands for action recognition from a bottom-up perspective~\cite{sun2018actor,zhang2023helping}. 

\begin{table}[t]
\centering
\resizebox{0.4\textwidth}{!}{
        \begin{tabular}{lccc}
            \hline
            Features  & VTCD  & VTCD + SAM~\cite{kirillov2023segment} \\
            \hline
            VideoMAE-SSL& 45.0  & 68.1 \\
            VideoMAE & 43.1 & 66.6 \\            
            InternVideo & 45.8 & 68.0 \\
        \end{tabular}}\vspace{-10pt}
    \caption{We apply VTCD to discover object tracking concepts and evaluate them with mIoU on the DAVIS'16 validation set.}
    \label{tab:VOS}\vspace{-15pt}
\end{table}
We validate the accuracy of the discovered object-centric concepts on the DAVIS'16 video-object segmentation (VOS) benchmark~\cite{perazzi2016benchmark}. To this end, we run VTCD on the training set and select concepts that have the highest overlap with the groundtruth tracks. 
We then use them to track new objects on the validation set and report results in Table~\ref{tab:VOS} (see appendix for implementation details). In this analysis, we solely evaluate representations that are not directly trained for VOS.
While solid performance is achieved by all representations, segmentation accuracy is limited by the low resolution of the concept masks. 
To mitigate this, we introduce a simple refinement step that samples random points inside the masks to prompt SAM~\cite{kirillov2023segment}. 
The resulting approach, shown as `VTCD + SAM', significantly improves performance. We anticipate that future developments in video representation learning will automatically lead to better VOS methods with the help of VTCD.
 
Finally, deeper layers contain concepts that build on object-centric representations to capture spatiotemporal events. For example, three models learn to identify containers before an object is dropped into them and two models track the object in the hand until it is dropped. One notable exception is InternVideo~\cite{wang2022internvideo}, which is primarily trained on images with limited spatiotemporal modeling. 
Perhaps surprisingly, these concepts are also found in the self-supervised VideoMAE~\cite{tong2022videomae}, which was never trained to reason about object-container relationships. This raises the question: Can intuitive physics models~\cite{chang2016compositional,wu2015galileo} be learned via large-scale video representation training? 


\vspace{-0.5em}
\section{Conclusion}
\vspace{-0.2em}
In this work, we introduced \AlgName, the first algorithm for concept discovery in video transformers. We empirically demonstrated that it is capable of extracting human-interpretable concepts from video understanding models and quantifying their importance for the final predictions. Using \AlgName, we discovered shared concepts among several models with varying objectives, 
revealing common processing patterns, such as a spatiotemporal basis in early layers. In later layers, useful, higher-level representations universally emerge, such as those responsible for object tracking.
Large-scale video representation learning is an active area of research at the moment~\cite{blattmann2023stable, bardes2024vjepa,sora} and our approach can serve as a key to unlocking its full potential.

\clearpage
\setcounter{page}{1}
\maketitlesupplementary

In this appendix, we report additional results, visualizations and implementation details.
We first conduct validation experiments for \AlgName to show that the discovered concepts align with human-interpretable groundtruth labels in Section~\ref{sec:abl}.
We then provide statistics of concepts importance distribution between layers in Section~\ref{sec:layer}.
Next, in Section~\ref{sec:head}, we provide further discussion and qualitative results showing how different concepts are captured in different self-attention heads from the same layer. 
We provide further implementation details in Section~\ref{sec:impl}. Finally, we discuss limitations of \AlgName in Section~\ref{sec:limitations}. Note that we include additional video results and corresponding discussions on the project web page. 

\section{Additional results}

\subsection{Concept validation}\label{sec:abl}
Directly measuring concept accuracy is not possible in an open world approach as we don't know a priori what concepts should be present in a model. There are, however, special cases where we can directly measure the accuracy of \textit{some} of the concepts. For example, TCOW is trained to track through occlusions, so we can expect to find concepts that correspond to the target object and containers/occluders in it. We perform this evaluation for VTCD and the random crop baseline~\cite{yeh2020completeness,fel2023craft,fel2023holistic} and report the mIoU between the best found concepts and the groundtruth masks in Table~\ref{tab:ValidationAblation} (top). These results validate the accuracy of VTCD, which is able to reach up to 94\% of the performance of the fully-supervised TCOW by discovering concepts in its intermediate representations.

\subsection{Quantitative analysis of per-layer concept importance} 
\label{sec:layer}
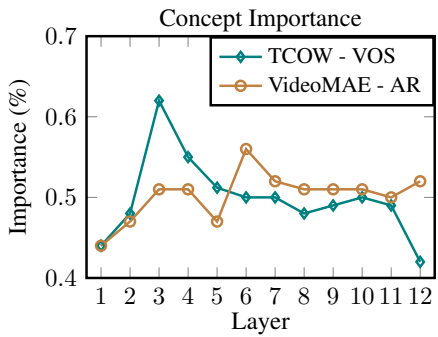
\begin{figure} [t]
	\begin{center}
     \centering 
\resizebox{0.35\textwidth}{!}{
\hspace{0.1cm}
\begin{tikzpicture} 
                 \begin{axis}[
                 line width=1.0,
                 title={Concept Importance},
                 title style={at={(axis description cs:0.5,1.09)},anchor=north,font=\normalsize},
                 xlabel={Layer},
                 ylabel={Importance (\%)},
                 xmin=0.5, xmax=12.5,
                 ymin=0.4, ymax=0.7,
                 xtick={1,2,3,4,5,6,7,8,9,10,11,12},
                 x tick label style={font=\normalsize, rotate=0, anchor=north},
                 y tick label style={font=\normalsize},
                 x label style={at={(axis description cs:0.5,0.05)},anchor=north,font=\normalsize},
                 y label style={at={(axis description cs:0,.5)},anchor=south,font=\at={(axis description cs:0.5,0.09)},anchor=north,font=\normalsize},
                 width=6.5cm,
                 height=5cm, 
                 ymajorgrids=false,
                 xmajorgrids=false,
                 major grid style={dotted,teal!20!black},
                 legend columns=1,
                 legend style={
                  nodes={scale=1, transform shape},
                  cells={anchor=west},
                  legend style={at={(1,1)},anchor=north east,row sep=0.01pt}, font =\small}
             ]
            \addlegendentry{TCOW - VOS}
            \addplot[line width=1pt, mark size=2pt, color=teal, mark=diamond]
                     coordinates {
(1,	0.44)
(2,	0.48)
(3,	0.62)
(4,	0.55)
(5,	0.512)
(6,	0.5)
(7,	0.5)
(8,	0.48)
(9,	0.49)
(10,0.5)
(11,0.49)
(12,0.42)
};

            \addlegendentry{VideoMAE - AR}
            \addplot[line width=1pt, mark size=2pt, color=brown, mark=o,error bars/.cd, y dir=both, y explicit,]
                     coordinates {
(1.00,	0.44)
(2.00,	0.47)
(3.00,	0.51)
(4.00,	0.51)
(5.00,	0.47)
(6.00,	0.56)
(7.00,	0.52)
(8.00,	0.51)
(9.00,	0.51)
(10.0,	0.51)
(11.0,	0.50)
(12.0,	0.52)
};  
              \end{axis}
\end{tikzpicture}}
\end{center}
\vspace{-0.6cm}
\caption{The average concept importance over all model layers for a VOS model (TCOW) and an action recognition model (VideoMAE). Interestingly, while VideoMAE encodes important concepts both in the middle and late in the model, TCOW encodes most important concepts at layer three and the least important in the final layer.
}\label{fig:LayerConceptImportance}
\vspace{-10pt}
\end{figure}

We now quantify the importance of each model layer for the two target models analyzed in Section~\ref{sec:GeneralConcepts} in the main paper. To this end, we calculate the average concept importance ranking per-layer and then normalize this value, which results in a $[0-1]$ score, where higher values indicate more important layers, and plot the results in Figure~\ref{fig:LayerConceptImportance}.

We immediately see similarities and differences between the two models. For example, the first two layers are less important than mid layers for both models. For VideoMAE, the middle (6) and end layer (12) are the most important. Interestingly, for TCOW, the most important layer by far is layer 3, while  the final layer is the least important. This makes intuitive sense since TCOW is an object tracking model, hence it most heavily utilizes spatiotemporal positional information and object-centric representations in early-to-mid layers. In contrast, VideoMAE is trained for action classification, which requires fine-grained, spatiotemporal concepts in the last layers.

\subsection{Uniqueness of head concepts} 
\label{sec:head}
\begin{figure}
    \centering   
\includegraphics[width=0.45\textwidth]{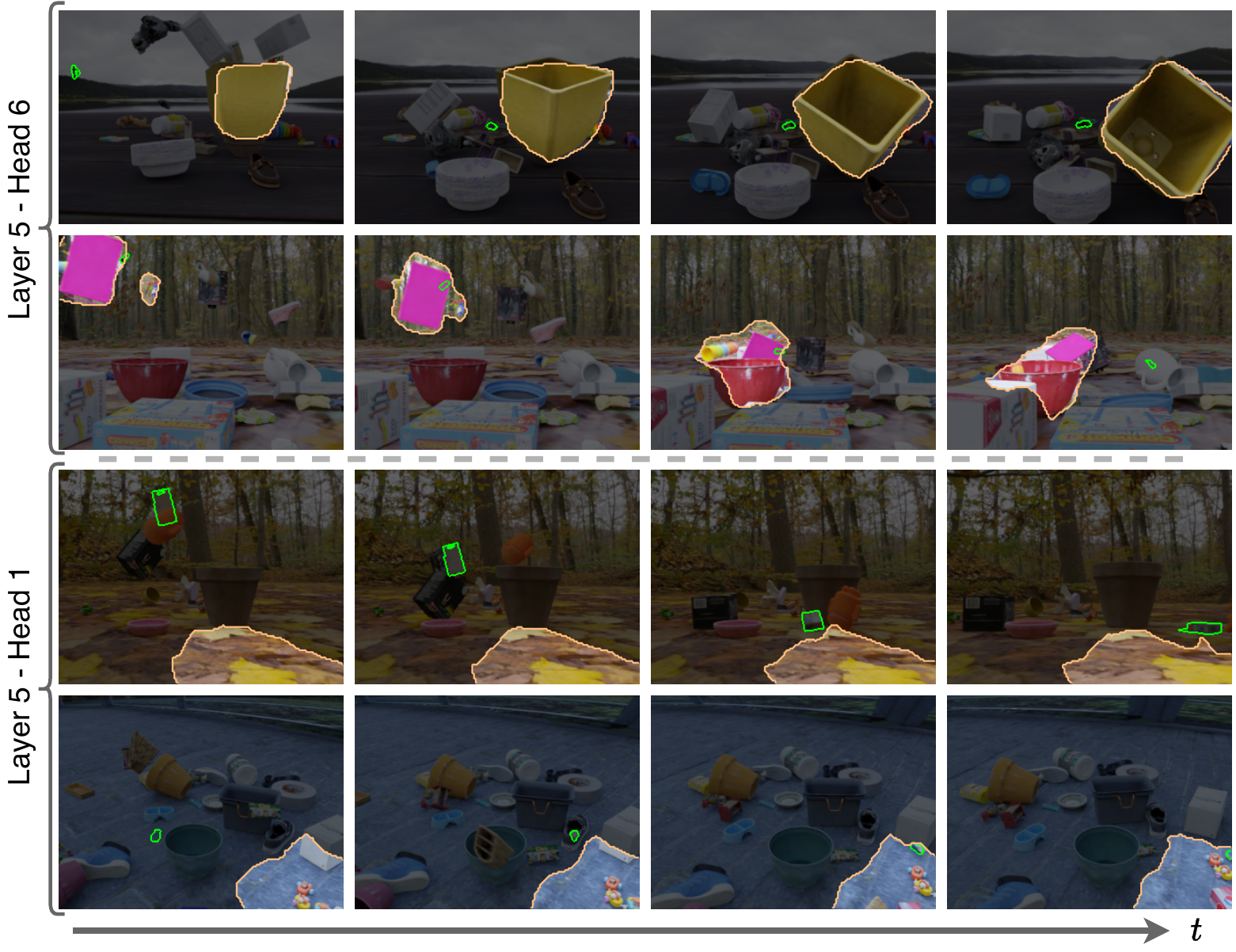}
    \caption{Different heads in the same layer encode concepts capturing different phenomena. In layer 5 of TCOW~\cite{van2023tracking}, head 6 (top two rows) highlights falling objects, while head 1 (bottom two rows) captures spatial position.}
    \label{fig:SameLayerDiffHead}
\end{figure}

As discussed in Section 4 in the main paper, we qualitatively visualize the concepts from the same layer but different heads of a model to demonstrate that the heads encode diverse concepts. For example, Figure~\ref{fig:SameLayerDiffHead} shows that discovered concepts in heads one and six in layer five of the TCOW~\cite{van2023tracking} model encode unrelated concepts (\eg positional and falling objects). This corroborates existing work~\cite{voita2019analyzing,amir2021deep,Karim_2023_CVPR,olsson2022context,elhage2021mathematical} that heads capture independent information and are therefore a necessary unit of study using \AlgName.

\begin{table}[t]
\centering
\resizebox{0.4\textwidth}{!}{
        \begin{tabular}{lccc}
            Method & Target & Occluders & Containers \\
            \hline
            Baseline & 3.0 & 31.5 & 43.3 \\
            VTCD (Ours) & \bf{19.2} & \bf{69.7} & \bf{73.8} \\
            \hline
            TCOW (supervised) & 36.8 & 76.8 & 78.2 \\
        \end{tabular}} 
    \caption{Evaluating the accuracy of object tracking concepts found in the TCOW model by VTCD and the random crop baseline used in other recent methods~\cite{yeh2020completeness,fel2023craft,fel2023holistic}.}
    \label{tab:ValidationAblation}
\end{table}

\section{Implementation details}
\label{sec:impl}
\smallsec{Concept discovery.} When generating tubelets (Section~\ref{sec:tubelets}), we use 12 segments and set all other hyperparameters to the Scikit-Image~\cite{van2014scikit} defaults, except for the compactness parameter, which is tuned on a held-out set for each model to the following values: TCOW - 0.01, VideoMAE - 0.1, SSL-VideoMAE - 0.1, InternVideo - 0.15. When clustering concepts using CNMF (Section~\ref{sec:ConceptClustering}) we follow the same protocol as~\cite{amir2021deep} and use the Elbow method, with the Silhouette metric~\cite{rousseeuw1987silhouettes} as the distance, to select the number of clusters with a threshold of 0.9. 

\smallsec{Concept importance.} For all importance rankings using CRIS, we use the original loss the models were trained with. For InternVideo, we use logit value for the target class by encoding the class name with the text encoder, and then taking the dot product between the text and video features. We use 4,000 masking iterations for all models, except for TCOW~\cite{van2023tracking}, where we empirically observe longer convergence times and use 8,000 masks.


\smallsec{Concept pruning with VTCD.} The six classes targeted in the concept pruning application (Table~\ref{tab:ApplicationResults} in the main paper) are as follows:

\begin{enumerate}
    \item Pouring something into something until it overflows
    \item Spilling something behind something
    \item Spilling something next to something
    \item Spilling something onto something
    \item Tipping something with something in it over, so something in it falls out
    \item Trying to pour something into something, but missing so it spills next to it
\end{enumerate}

\smallsec{Semi-Supervised VOS with VTCD.} To evaluate VOS performance on the DAVIS'16 benchmark~\cite{perazzi2016benchmark} with any pretrained video transformer, we first identify self-attention heads that encode object-centric concepts on the training set. Specifically, we calculate the mIoU between every concept found by VTCD (\ie the set of tubelets belonging to that concept) and the groundtruth labels for each training video. We then record the heads in which the best performing concepts came from. Next, we run \AlgName on the validation set using only the heads containing object-centric concepts. To select the final concept for evaluation, we choose the one with the highest mIoU with the first frame label (\ie the query mask). 

To generate tubelets for an entire video, we simply use non-overlapping sliding windows and run SLIC on the temporally concatenated features. We also leverage SAM~\cite{kirillov2023segment} for post-processing the tubelets generated by \AlgName. Note that SAM can take as input a set of points, a bounding box, or both. For each frame and corresponding mask from the \AlgName tubelet, we generate the smallest bounding box surrounding the mask. We also calculate the centroid of the mask and then sample two points from a Gaussian centered at the centroid, with covariance $(L/10, W/10)$ where $L$ and $W$ are the length and width of the mask, respectively. We then pass both the box and points to SAM ViT-h to produce the post-processed mask for that frame.

\section{Limitations}
\label{sec:limitations}
One limitation of our method is the need to manually set the SLIC compactness hyper-parameter. Additionally, the compactness property of SLIC makes it challenging to capture concepts that are not spatially localized. Finally, there is a high computational requirement for computing the Rosetta score in Equations~\ref{eq:r} and~\ref{eq:r2} as $D$ grows.

\vspace{5pt}
{\small
\noindent{\bf Acknowledgements.} We acknowledge financial support from the Canadian NSERC Discovery Grants. K.G.D.\ contributed to this work in their personal capacity as Associate Professor at York University. Thanks to Greg Shakhnarovich for feedback on the paper. 
}

\clearpage
\newpage

{
    \small
    \bibliographystyle{ieeenat_fullname}
    \bibliography{main}
}


\end{document}